 \documentclass[pmlr,twocolumn,10pt]{jmlr} 

\usepackage{booktabs}
\usepackage{placeins} 
\usepackage{flafter}   
\usepackage{float}     
\usepackage{stfloats}  
\usepackage{mathtools}
\usepackage{siunitx}

\usepackage[switch]{lineno}

\newcommand{\Prb}{\mathbb{P}}
\newcommand{\E}{\mathbb{E}}

\newcommand{\equal}[1]{{\hypersetup{linkcolor=black}\thanks{#1}}}

\theorembodyfont{\upshape}
\theoremheaderfont{\scshape}
\theorempostheader{:}
\theoremsep{\newline}

 \title[Robust Ensemble]{Provably Robust Pre-Trained Ensembles for Biomarker-Based Cancer Classification}

\author{%
\Name{Chongmin Lee}
\Email{chongminlee@hsph.harvard.edu}\\
\addr Department of Biostatistics, Harvard University
\AND
\Name{Jihie Kim}\equal{Corresponding author.}
\Email{jihie.kim@dgu.edu}\\
\addr Department of Computer Science, Dongguk University
}

\begin{document}

\maketitle

\begin{abstract}
Certain cancer types, notably pancreatic cancer, are difficult to detect at an early stage, motivating robust biomarker-based screening. Liquid biopsies enable non-invasive monitoring of circulating biomarkers, but typical machine learning pipelines for high-dimensional tabular data (e.g., random forests, SVMs) rely on expensive hyperparameter tuning and can be brittle under class imbalance. We leverage a meta-trained Hyperfast model for classifying cancer, accomplishing the highest AUC of 0.9929 and simultaneously achieving robustness especially on highly imbalanced datasets compared to other ML algorithms in several binary classification tasks (e.g. breast invasive carcinoma; BRCA vs. non-BRCA). We also propose a novel ensemble model combining pre-trained Hyperfast model, XGBoost, and LightGBM for multi-class classification tasks, achieving an incremental increase in accuracy (0.9464) while merely using 500 PCA features; distinguishable from previous studies where they used more than 2,000 features for similar results. Crucially, we \emph{demonstrate robustness under class imbalance}: empirically via balanced accuracy and minority-class recall across cancer-vs.-noncancer and cancer-vs.-rest settings, and theoretically by showing (i) a prototype-form final layer for Hyperfast that yields prior-insensitive decisions under bounded bias, and (ii) minority-error reductions for majority vote under mild error diversity. Together, these results indicate that pre-trained tabular models and simple ensembling can deliver state-of-the-art accuracy and improved minority-class performance with far fewer features and no additional tuning.
\end{abstract}

\paragraph*{Data and Code Availability}
All primary tissue datasets used in this work are publicly available via The Cancer Genome Atlas Program (TCGA).
External liquid-biopsy RNA cohorts are available from GEO under accessions GSE68086 and GSE71008.
The training code as well as the pipeline for reproducing the results can be found
\href{https://github.com/ChongminLee/hyperfast_ensemble}{here}.

\paragraph*{Institutional Review Board (IRB)}
This work does not require IRB approval.

\section{Introduction}
\label{sec:intro}

Early detection of cancer has always been a challenge for researchers and medical practitioners due to its complexity underlying in each patient and numerous causes that lead to cancer. Thus, several molecular features such as patients’ gene expression profiles or histopathological features, namely cell morphology are leveraged to classify and diagnose cancer. Recently, imaging features such as MRI, CT scans, and PET scans analyzed in \citet{mallick2019brain,gupta2023deep} are often used for detecting cancer and further analyzing the cancer type. However, some prevailing methods have limits that it is time-consuming \citep{isin2016review} or often expensive to conduct. Biomarkers, on the other hand, are relatively easy to attain from liquid biopsy, a non-invasive method. Thus, in this paper we develop predictive models (i.e. statistical and machine learning algorithms) to exploit these biomarker levels to infer cancer and classify them more accurately.

The adoption of such predictive models is useful in wide applications to the real-world problems. Predictive models aid in treatment for patients with cancer and providing early diagnosis and prognosis of cancer. Additionally, these models can be used for cross-validation of other prediction methods exploiting genetic features such as chromosomal rearrangements or DNA sequence. We specifically delve into the causal relationship between biomarkers and cancer types in this paper, using 40,000+ biomarkers and 33 types of cancer.

For the development of the predictive model, we use several machine learning algorithms for classification of tabular data. Classification of cancer through machine learning techniques have been studied extensively. \citet{zelli2023classification, michael2022optimized} propose that some boosting algorithms such as XGBoost and LightGBM perform cancer-type classification very effectively. However, we explore how the meta-trained Hyperfast model \citep{bonet2024hyperfast} performs classification with the best accuracy on certain binary classification tasks. Hyperfast is a meta-trained model with several genetic/genomic datasets included in meta-training/testing/validation datasets namely UK Biobank \citep{sudlow2015uk} and HapMap3 \citep{consortium2010hapmap}, enabling powerful classification even on high-dimensional biomedical data including our TCGA dataset. We also propose a novel ensemble model combining Hyperfast with two boosting algorithms for multiclass classification. Similar studies have been proposed where one explores the ensemble classification of cancer types and biomarker identification \citep{wang2022ensemble}, and the other leverages ensemble learning for biomarker-based early detection of pancreatic cancer \citep{nene2023serum}. Additionally, regarding evaluating the robustness of a pre-trained model, \citet{pmlr-v97-hendrycks19a} show that even when standard accuracy gains are small, pre-training markedly improves robustness (to label corruption and class imbalance), adversarial accuracy, out-of-distribution detection, and calibration, reinforcing our choice of a pre-trained backbone and hard-voting ensemble for imbalanced biomarker classification.

Our contribution is that we use Hyperfast, a pre-trained model that eliminates the need for additional hyperparameter tuning prior to training and testing, reducing the time for classifying cancer. Additionally, our ensemble model for multiclass classification achieved the SOTA-level performance while leveraging less than 1.5\% of the available biomarker features (i.e., 500 PCA features). Previous results show they leveraged 2,000 gene expression features for attaining similar results on cancer classification \citep{deng2022hybrid}. Showing that predicting cancer can be performed with a very low number of biomarkers could impact future designs of classification models and decrease the cost of the overall task \citep{bartusiak2022predicting}. Next, we discovered that the pre-trained Hyperfast model performs especially robust with binary classification on highly imbalanced datasets (e.g., Breast invasive carcinoma vs. non-BRCA). In this paper, we rigorously delve into the background as to why our ensemble model performs robust on highly imbalanced data. Finally, we were able to verify that Hyperfast model performs classification well even on large datasets with $>$ 1,000 training examples and $>$ 10 target classes, which is crucial for further analysis on high-dimensional biomedical data.

\section{Methods}
\label{sec:methods}
\subsection{Dataset}

Humans have numerous different biomarkers including but not limited to genomic biomarkers \citep{chaudhari2016identification} and transcriptomic biomarkers \citep{zhang2012development} such as mRNA levels. Our dataset, attained from TCGA (The Cancer Genome Atlas Program) consists of 40,000+ biomarker features for 16,784 different patients diagnosed with 33 types of cancer and 1 cancer negative. In other words, each data sample is a row of 40,000+ biomarkers represented as numerical values \citep{hashim2022subomiembed}. 

\begin{figure}[t]
\label{fig:tsne}
\floatconts
  {fig:tsne}
  {\caption{t-Distributed Stochastic Neighbor Embedding (t-SNE) of the biomarker features in the dataset. Results are color-coded according to cancer type, which the labels can be found in Table 1}}
  {\includegraphics[width=\columnwidth]{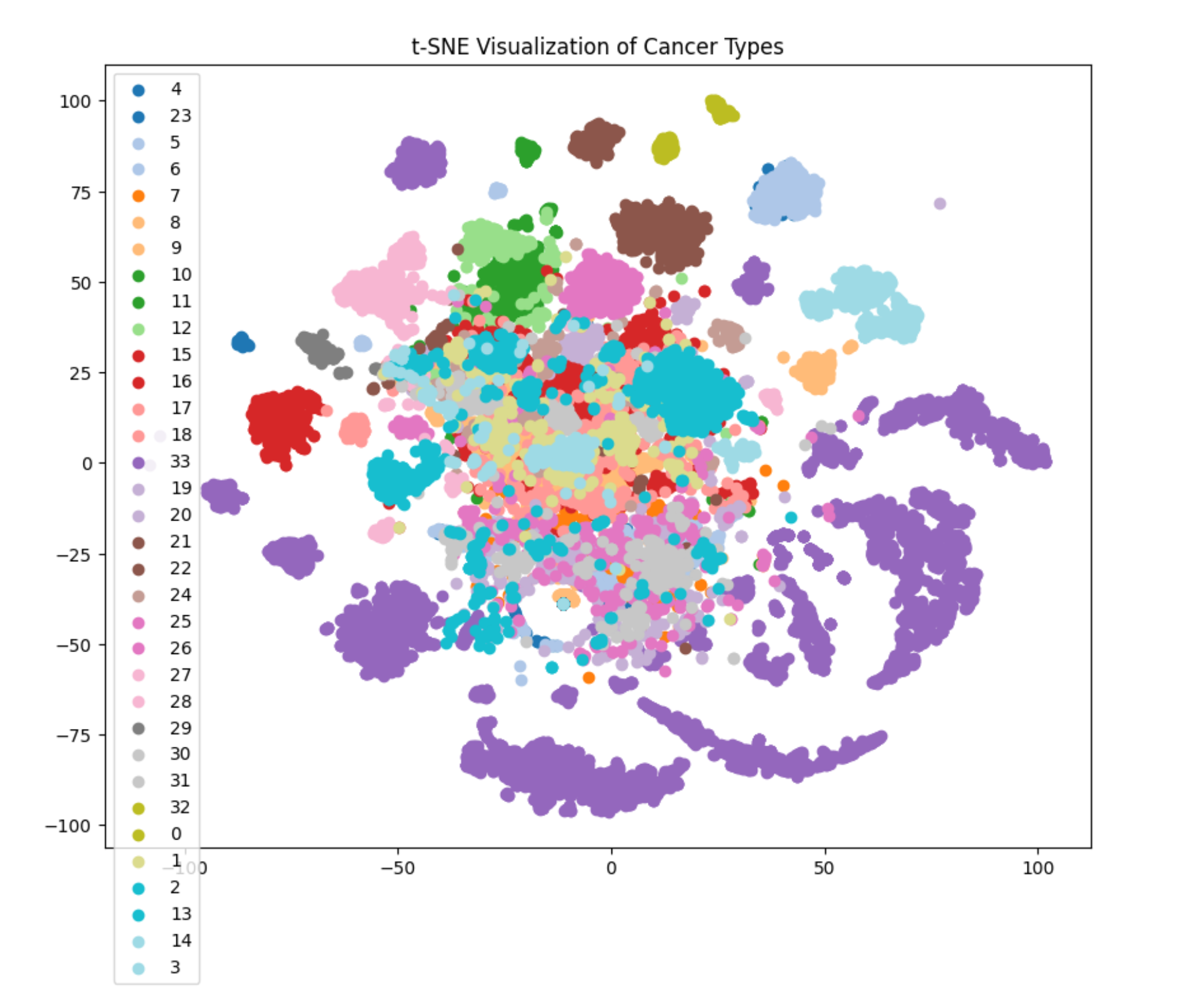}}
\end{figure}

Because our dataset includes many cancer types, we visualize the dataset to understand higher-level relationships of cancer with biomarkers. Prior to t-SNE visualization, we perform dimension reduction using PCA by selecting 200 biomarker features (\citet{devassy2020dimensionality, song2019improved}): visualizing different cancer slightly more distinctly. \figureref{fig:tsne} shows the distribution of samples in our dataset based on cancer type and illustrates the imbalanced trait of our data, with some cancer containing significantly more samples than others. For instance, breast invasive carcinoma (BRCA) includes 1,084 samples, while there are only 36 samples of cholangiocarcinoma (CHOL). 

\figureref{fig:tsne} provides a t-SNE visualization of the high-dimensional biomarker features of various cancer types, reduced to a two-dimensional space. This visualization helps to identify potential patterns and groupings in the data. While some cancer types such as Bladder Urothelia Carcinoma (BLCA) and Uterine Corpus Endometrial Carcinoma (UCEC) appear to overlap in the t-SNE plot, this should not necessarily be interpreted as an inherent difficulty in classification. The distinct separation of Uveal Melanoma (UVM) from Cholangiocarcinoma (CHOL) and the clear segregation of 'Normal (Cancer Negative)' samples highlight the local structure preserved by t-SNE, yet these observations are exploratory rather than conclusive.

\tableref{tab:cancer-labels} also shows the imbalanced nature of our dataset. To further perform multi-class classification on each cancer, we perform label encoding before preprocessing the data so that each cancer type is encoded as numerical value, enabling us to perform classification techniques using Python \citep{jia2023multi}.

For all TCGA experiments, we form a single stratified 75/25 train--test split that preserves class proportions across splits. PCA is always fit on the training split and then applied to the test split, and all models and hyperparameters are selected using training data only. The headline binary and multiclass metrics in Section~\ref{sec:results} are reported on this held-out test split, which prevents any train--test leakage.

\begin{table}[t] 
\label{tabl:results}
\floatconts
  {tab:cancer-labels}
  {\caption{Cancer Type Acronyms and Corresponding Counts (descending) with Label Encodings}}
  {%
    \setlength{\tabcolsep}{6pt}\footnotesize 
    \begin{tabular}{@{}lrr@{}}
      \toprule
      \bfseries Cancer\_Type\_Acronym & \bfseries Count & \bfseries Label\_Encoded\\
      \midrule
      Normal (Cancer Negative) & 5,817 & 33\\
      BRCA & 1,084 & 2\\
      GBM & 592 & 8\\
      OV & 585 & 19\\
      LUAD & 566 & 16\\
      UCEC & 529 & 30\\
      HNSC & 523 & 9\\
      LGG & 514 & 14\\
      KIRC & 512 & 11\\
      THCA & 500 & 28\\
      PRAD & 494 & 22\\
      LUSC & 487 & 17\\
      SKCM & 448 & 25\\
      STAD & 440 & 26\\
      COAD & 439 & 5\\
      BLCA & 411 & 1\\
      LIHC & 372 & 15\\
      CESC & 297 & 3\\
      KIRP & 283 & 12\\
      SARC & 255 & 24\\
      LAML & 200 & 13\\
      PAAD & 184 & 20\\
      ESCA & 182 & 7\\
      PCPG & 178 & 21\\
      READ & 155 & 23\\
      TGCT & 149 & 27\\
      THYM & 123 & 29\\
      ACC & 92 & 0\\
      MESO & 87 & 18\\
      UVM & 80 & 32\\
      KICH & 65 & 10\\
      UCS & 57 & 31\\
      DLBC & 48 & 6\\
      CHOL & 36 & 4\\
      \bottomrule
    \end{tabular}
  }%
\end{table}

\subsection{Algorithm of Hyperfast Model}
\label{sec:hyperfast}

\paragraph{Setup and meta-objective.}
Let $\mathcal{X}$ be the input space, $\mathcal{Y}=\{1,\dots,K\}$ the label set,
and let a task $t$ draw a \emph{support} set $S_t=\{(x_i,y_i)\}_{i=1}^{n_S}$ and a \emph{query}
set $Q_t=\{(x_j,y_j)\}_{j=1}^{n_Q}$. A hypernetwork $h_{\phi}$ maps the support to
main-network weights $\theta^* = h_{\phi}(S_t)$. Denote the feature map produced by the
(main) network with weights $\theta^*$ as
\[
\Phi_{S_t}(x) = \Phi\!\big(x;\theta^*\big)\in\mathbb{R}^d .
\]
The (per-task) loss is
\begin{equation}
\label{eq:meta-loss}
\mathcal{L}_t(\phi)\;=\;\frac{1}{|Q_t|}\sum_{(x,y)\in Q_t}\!
\ell\!\Big(f_{\theta^*}(x),\,y\Big)
\quad\text{with}\quad \theta^*=h_{\phi}(S_t),
\end{equation}
and the meta-training objective minimizes $\sum_{t\in\mathcal{T}_{\text{meta-train}}}\mathcal{L}_t(\phi)$.

\paragraph{Prototype form of the final layer.}
Write $S_{t,c}=\{(x_i,y_i)\in S_t:\,y_i=c\}$ and define the class
\emph{prototype}
\begin{equation}
\label{eq:prototype}
\mu_c(S_t)\;\coloneqq\;\frac{1}{|S_{t,c}|}\sum_{(x_i,y_i)\in S_{t,c}}\!\Phi_{S_t}(x_i).
\end{equation}
As shown in \lemmaref{lem:prototype-form}, the Hyperfast final affine layer can be
chosen (up to a global positive scale $\alpha$) so that its class weight
vectors are proportional to the prototypes and the bias is support-dependent:
\begin{equation}
\label{eq:hf-decision}
\hat y(x;S_t)
=\operatorname*{arg\,max}_{c\in\mathcal{Y}}
\Big\langle \mu_c(S_t),\,\Phi_{S_t}(x)\Big\rangle + \tilde b_c(S_t).
\end{equation}
Under the bounded-embedding assumption $\|\Phi_{S_t}(x)\|\le B$ and a $\rho$-Lipschitz
dependence of the hypernetwork bias on $S_t$, the relative bias is uniformly controlled:
\begin{equation}
\label{eq:bias-bound}
\big|\tilde b_{c'}(S_t)-\tilde b_c(S_t)\big|\;\le\;2\rho B
\qquad\forall\,c,c'\in\mathcal{Y}.
\end{equation}
Equations~\eqref{eq:hf-decision}–\eqref{eq:bias-bound} are the forms used in our
experiments. (In \sectionref{sec:results} we will appeal to \theoremref{thm:HFmargin}
to explain the observed prior-insensitivity on imbalanced tasks and to Proposition~10
for the prior-shift case.)

\begin{algorithm2e}
\caption{HyperFast Algorithm}
\label{alg:hyperfast-2e}
\DontPrintSemicolon
\KwIn{Support set $S_t$, number of layers $L$}
\KwOut{Predictive distribution $p_\theta(y\mid x,S)$}
$h(S_t)\leftarrow \theta^{*}$\;
$\theta^{*} \leftarrow \textsc{HyperFast}(S_t)$\;
$\theta_{\text{main}} \leftarrow \{\theta_{\text{main},\ell}\}_{\ell=1}^{L}$\;
$\textit{transformed\_S} \leftarrow \textsc{initial\_transform}(S_t)$\;
\For{$\ell\leftarrow 1$ \KwTo $L$}{
  $\theta_{\text{main},\ell} \leftarrow \textsc{predict\_weights}(\textit{transformed\_S})$\;
}
Minimize $\displaystyle \mathcal{L}_{t}
= \sum_{t\in \mathcal{T}_{\text{meta-train}}} \mathcal{L}_t\!\left(f_{\theta^{*}}(S_t,x)\right)$\;
\Return $p_{\theta}(y\mid x,S) \leftarrow f_{\theta^{*}}(S)(x)$\;
\end{algorithm2e}

\subsection{Ensemble Model}

Prior to inputting the data into the ensemble model, we first preprocess the TCGA data by dropping missing values and standardizing for Principal Component Analysis. We conduct PCA to reduce the dimensionality of the dataset as well as to monitor how accuracy changes as the number of features varies (e.g., 200, 500). This step is crucial since by monitoring how performance changes as the number of Principal Components (PCs) used differs, we can distinguish the models based on their robustness when more than 10,000 PCs are utilized for model training. Further analysis will be presented in \sectionref{sec:results}.

\begin{algorithm2e}
\caption{Ensemble Model with PCA}
\label{alg:ensemble-pca-2e}
\DontPrintSemicolon
\SetKwFunction{EnsemblePredict}{EnsemblePredict}
\SetKwFunction{HyperFastPredict}{HyperFastPredict}
\SetKwFunction{XGBoostPredict}{XGBoostPredict}
\SetKwFunction{LightGBMPredict}{LightGBMPredict}
\SetKwFunction{MajorityVote}{MajorityVote}
\SetKwProg{Proc}{Procedure}{}{}
\SetKwProg{Fn}{Function}{}{}

\Proc{\EnsemblePredict{$X$}}{
  $X_{\text{scaled}}\leftarrow\text{Standardize for PCA}(X)$\;
  $X_{\text{reduced}}\leftarrow \mathrm{PCA}(\mathrm{fit\_transform}(X_{\text{scaled}}))$\;
  $\widehat{y}_{\mathrm{hf}}\leftarrow \HyperFastPredict(X_{\text{reduced}})$\;
  $\widehat{y}_{\mathrm{xgb}}\leftarrow \XGBoostPredict(X_{\text{reduced}})$\;
  $\widehat{y}_{\mathrm{lgbm}}\leftarrow \LightGBMPredict(X_{\text{reduced}})$\;
  $\textit{final\_pred}\leftarrow \MajorityVote([\widehat{y}_{\mathrm{hf}},\widehat{y}_{\mathrm{xgb}},\widehat{y}_{\mathrm{lgbm}}])$\;
  \Return \textit{final\_pred}\;
}

\BlankLine
\Fn{\HyperFastPredict{$X$}}{
  \Return $f_{\theta^\ast}(S)(x)$\;
}

\BlankLine
\Fn{\XGBoostPredict{$X$}}{
  Minimize $\displaystyle
  \mathcal{L}_{M}=\sum_{j}\left[\left(\sum_{i\in I_j} g_i\right) w_j
  +\frac{1}{2}\left(\sum_{i\in I_j} h_i + \lambda\right) w_j^2\right]+\gamma T$\;
  \Return predictions\;
}

\BlankLine
\Fn{\LightGBMPredict{$X$}}{
  \eIf{binary classification}{
    Minimize $\displaystyle
      \mathcal{L}(y,f(x))
      = -\frac{1}{N}\sum_{i=1}^{N}\big[y_i\log(p_i)+(1-y_i)\log(1-p_i)\big]$\;
  }{
    Minimize $\displaystyle
      \mathcal{L}(y,f(x))
      = -\frac{1}{N}\sum_{i=1}^{N}\sum_{k=1}^{K} y_{i,k}\log(p_{i,k})$\;
  }
  \Return predictions\;
}

\BlankLine
\Fn{\MajorityVote{predictions}}{
  \Return $\displaystyle \arg\max_{c}\sum_{i=1}^{n}\mathbf{1}(\hat{y}_i=c)$\;
}
\end{algorithm2e}

We propose an ensemble model combining pre-trained Hyperfast, XGBoost, and LightGBM to conduct multi-class classification tasks to identify 34 different target classes. We compare the performance results with other several classification techniques as the baseline. For the baseline, we use XGBoost \citep{dimitrakopoulos2018pathway}; Gradient Boosting \citep{kwak2017multi, kumar2024analysis}; Random Forest Classifier \citep{breiman2001random}; LightGBM \citep{wang2017lightgbm}; Logistic Regression \citep{nick2007logistic}; kNN \citep{guo2003knn}; linear and non-linear Support Vector Machines (SVMs)\citep{nie2020decision}; Gaussian Naive Bayes \citep{jahromi2017nonparametric}; and Hyperfast \citep{bonet2024hyperfast}.

We choose XGBoost and LightGBM for our ensemble model due to their high performance in multi-class classification. For instance, XGBoost and LightGBM achieved an accuracy of 0.9399 and 0.9411 respectively leveraging 500 PCs, while other classification techniques such as kNN had an accuracy of 0.8832. Additionally, the objective function of both boosting algorithms is \(\sum_{i=1}^{n} \mathcal{L}(y_i,\hat{y}_i) + \sum_{k=1}^{K} \sigma(f_k)\), comprised of two parts: the loss function that measures how well the model fits the training data, and a regularization term that penalizes the complexity of the model \citep{sheng2022optimized}. The regularization term can be written as $\sigma(f_k)=\gamma T + \frac{1}{2}\lambda \sum_{j=1}^{T} w_j^2$, where $T$ refers to the number of leaves in the decision tree, which represents prediction or outcome of the decision process. With both frameworks using similar forms of regularization to avoid overfitting, namely L1(Lasso) and L2(Ridge), they will both achieve high accuracy if the regularization is successfully conducted. However, since both boosting algorithms are held with the risk of overfitting if not properly regularized, we ensemble the Hyperfast model to resolve the gap. In real-world biomedical applications, numerous tabular datasets manifest very high dimensionality and class imbalance, making boosting algorithms susceptible to overfitting. Thus, by hard voting Hyperfast with the highly accurate two gradient-boosted trees, we can achieve both accuracy and robustness especially in high dimensional biomedical datasets. 

Now, the preprocessed data is inserted into the model where each classifier returns the prediction of the specific cancer type based on biomarker input given. The final prediction is returned by majority voting the outputs of the classifiers which will be used for further calculating accuracy and balanced accuracy \citep{chawla2010imbalanced} (a powerful metric for tasks on imbalanced dataset).

\paragraph{Base learners and posteriors.}
We form an ensemble of $M=3$ base classifiers:
$m=1$ (Hyperfast), $m=2$ (XGBoost), and $m=3$ (LightGBM).
Each learner outputs a posterior $\pi^{(m)}(c\mid x)$ and a label prediction
$g_m(x)=\arg\max_{c}\pi^{(m)}(c\mid x)$.

\paragraph{Hard vote rule.}
The ensemble prediction is the (three-way) hard majority vote
\begin{equation}
\label{eq:mv}
\hat y_{\text{ens}}(x)\;=\;
\operatorname*{arg\,max}_{c\in\mathcal{Y}}
\sum_{m=1}^{3}\mathbf{1}\!\big\{g_m(x)=c\big\},
\end{equation}
with ties broken by the largest mean posterior
$\arg\max_{c}\frac{1}{3}\sum_{m}\pi^{(m)}(c\mid x)$.
All base models are trained on the same PCA-reduced features ;
the boosting objectives are the standard regularized risks discussed below.

\paragraph{Why it helps on minority classes.}
In \sectionref{sec:results}–\sectionref{sec:conclusion} we verify empirically that
the hard vote~\eqref{eq:mv} reduces minority-class error when base errors are
diverse; this aligns with Proposition~7 and \corollaryref{cor:indep} (minority-error
drop under mild error-independence). We also discuss robustness of Hyperfast under prior shift using Proposition~10 and \theoremref{thm:HFmargin}.

\section{Results}
\label{sec:results}

Overall, we have over 40,000 biomarker features to analyze and leverage to classify cancer. However, one of our goals is to find how few biomarker features are necessary to classify cancer while achieving comparable accuracy to the SOTA performance, leveraging multi-omics data for TCGA cancer type classification \citep{hashim2022subomiembed}. Thus, we conduct all our experiments for different number of PCs (i.e. number of biomarker features). Let x represent the number of PCs, where $x \in {200, 500, 1000, 2000, 5000, 10000, 15000}$ for multi-class classification and $x \in {200, 500, 1000, 2000}$ for binary classification. Since even using 1,000 biomarker features achieved over 0.99 accuracy for binary classification, we only conducted the experiment until 2,000 PCs.

\subsection{Binary Classification}

We first conduct binary classification on cancer vs. non-cancer (i.e., Normal) task and calculate the results in six different metrics: Accuracy, AUC, F1-score, PPV, Sensitivity, and Specificity. \tableref{tab:results-cancer} indicates that the meta-trained Hyperfast model performs the best along with XGBoost achieving an accuracy of 0.9971 when using 200 PCs. However, as number of PCs increases, the accuracy as well as other metrics such as AUC of the other two boosting algorithms surpass the accuracy and AUC of Hyperfast. Nevertheless, Hyperfast performed comparably to XGBoost and LightGBM, achieving AUC over 0.98 in classification using 1,000 and 2,000 PCs. These results show how boosting algorithms outperform the pre-trained model on classification of mildly imbalanced dataset (i.e., Cancer: 5,084 samples vs. non-Cancer: 11,700 samples). 

\begin{table}[t]
\floatconts
  {tab:results-cancer}
  {\caption{Binary: Cancer vs. Non-Cancer Classification Results}}
  {%
    \sisetup{table-number-alignment=center}
    \setlength{\tabcolsep}{4pt}
    \resizebox{\columnwidth}{!}{%
      \begin{tabular}{>{\centering\arraybackslash}p{4cm}cccccc}
      \hline
      \textbf{Model (\# of PCs)} & \textbf{Accuracy} & \textbf{AUC} & \textbf{F1-Score} & \textbf{PPV} & \textbf{Sensitivity} & \textbf{Specificity} \\ \hline
      \textbf{Hyperfast (200)} & \textbf{0.9971} & 0.9590 & 0.9269 & 0.8664 & \textbf{0.9965} & 0.9215 \\ 
      \textbf{Hyperfast (500)} & 0.9380 & 0.9454 & 0.9136 & 0.8647 & 0.9683 & 0.9226 \\ 
      \textbf{Hyperfast (1,000)} & 0.9909 & 0.9866 & 0.9864 & \textbf{1.0000} & 0.9732 & \textbf{1.0000} \\ 
      \textbf{Hyperfast (2,000)} & 0.9886 & 0.9834 & 0.9828 & 0.9985 & 0.9676 & 0.9993 \\ \hline
      \textbf{XGBoost (200)} & \textbf{0.9971} & \textbf{0.9958} & \textbf{0.9958} & \textbf{1.0000} & 0.9915 & \textbf{1.0000} \\ 
      \textbf{XGBoost (500)}  & \textbf{0.9967} & \textbf{0.9951} & \textbf{0.9950} & \textbf{1.0000} & \textbf{0.9901} & \textbf{1.0000} \\ 
      \textbf{XGBoost (1,000)} & \textbf{0.9971} & \textbf{0.9958} & \textbf{0.9958} & \textbf{1.0000} & \textbf{0.9915} & \textbf{1.0000} \\ 
      \textbf{XGBoost (2,000)} & \textbf{0.9971} & \textbf{0.9958} & \textbf{0.9958} & \textbf{1.0000} & \textbf{0.9915} & \textbf{1.0000} \\ \hline
      \textbf{LightGBM (200)} & 0.9969 & 0.9954 & 0.9954 & \textbf{1.0000} & 0.9908 & \textbf{1.0000} \\ 
      \textbf{LightGBM (500)} & \textbf{0.9967} & \textbf{0.9951} & \textbf{0.9950} & \textbf{1.0000} & \textbf{0.9901} & \textbf{1.0000} \\ 
      \textbf{LightGBM (1,000)} & 0.9969 & 0.9954 & 0.9954 & \textbf{1.0000} & 0.9908 & \textbf{1.0000} \\ 
      \textbf{LightGBM (2,000)} & 0.9967 & 0.9951 & 0.9950 & \textbf{1.0000} & 0.9901 & \textbf{1.0000} \\ \hline
      \end{tabular}
    }
  }
  {}
\end{table}

\begin{table}[t]
\floatconts
  {tab:results-breast}
  {\caption{Binary: Breast Cancer Classification Results}}
  {%
    \sisetup{table-number-alignment=center}
    \setlength{\tabcolsep}{4pt}
    \resizebox{\columnwidth}{!}{%
      \begin{tabular}{>{\centering\arraybackslash}p{4cm}cccccc}
\hline
\textbf{Model (\# of PCs)} & \textbf{Accuracy} & \textbf{AUC} & \textbf{F1-Score} & \textbf{PPV} & \textbf{Sensitivity} & \textbf{Specificity} \\ \hline
\textbf{Hyperfast (200)} & \textbf{0.9988} & \textbf{0.9911} & \textbf{0.9910} & \textbf{1.0000} & \textbf{0.9822} & \textbf{1.0000} \\ 
\textbf{Hyperfast (500)} & \textbf{0.9990} & \textbf{0.9929} & \textbf{0.9928} & \textbf{1.0000} & \textbf{0.9858} & \textbf{1.0000} \\ 
\textbf{Hyperfast (1,000)} & \textbf{0.9981} & \textbf{0.9858} & \textbf{0.9856} & \textbf{1.0000} & \textbf{0.9715} & \textbf{1.0000} \\ 
\textbf{Hyperfast (2,000)} & 0.9800 & 0.8505 & 0.8243 & \textbf{1.0000} & 0.7011 & \textbf{1.0000} \\ \hline
\textbf{XGBoost (200)} & 0.9952 & 0.9644 & 0.9631 & \textbf{1.0000} & 0.9288 & \textbf{1.0000} \\ 
\textbf{XGBoost (500)} & 0.9964 & 0.9733 & 0.9726 & \textbf{1.0000} & 0.9466 & \textbf{1.0000} \\ 
\textbf{XGBoost (1,000)} & 0.9967 & 0.9751 & 0.9745 & \textbf{1.0000} & 0.9502 & \textbf{1.0000} \\ 
\textbf{XGBoost (2,000)} & 0.9957 & 0.9680 & 0.9669 & \textbf{1.0000} & 0.9359 & \textbf{1.0000} \\ \hline
\textbf{LightGBM (200)} & 0.9971 & 0.9786 & 0.9782 & \textbf{1.0000} & 0.9573 & \textbf{1.0000} \\ 
\textbf{LightGBM (500)} & 0.9979 & 0.9840 & 0.9837 & \textbf{1.0000} & 0.9680 & \textbf{1.0000} \\ 
\textbf{LightGBM (1,000)} & 0.9976 & 0.9822 & 0.9819 & \textbf{1.0000} & 0.9644 & \textbf{1.0000} \\ 
\textbf{LightGBM (2,000)} & \textbf{0.9974} & \textbf{0.9804} & \textbf{0.9800} & \textbf{1.0000} & \textbf{0.9609} & \textbf{1.0000} \\ \hline
      \end{tabular}
    }
  }
  {}
\end{table}

Next, we conduct binary classification identically using the three classification algorithms on significantly imbalanced dataset (e.g., LUAD: 566 samples vs. non-LUAD: 16,218 samples). Out of 33 types of cancer, we choose Breast Invasive Carcinoma (BRCA) and Lung adenocarcinoma (LUAD) for the target classes since they are one of the most common cancer types that elderly over 65 are diagnosed \citep{sung2021global}. \tableref{tab:results-breast} indicates that the Hyperfast model achieves the highest performance in all metrics including Accuracy when using equal or less than 1,000 PCs. However, the sensitivity of the Hyperfast model drops drastically when using 2,000 PCs, implying the possibility of feature redundancy or inclusion of noise \citep{xie2021artificial}. To further investigate this drop in sensitivity, we performed a series of additional analyses. We conducted a feature importance analysis using the sum of variances of PCs, which revealed that many of the additional PCs beyond the initial 1,000 did not significantly contribute to the model's predictions (variance $<$ 0.0002). Lower variance in the additional PCs suggests that these components are less informative and may introduce redundancy into the model. This redundancy can dilute the model's ability to identify relevant patterns, leading to a plateau or degradation in performance. Our cross-validation results indicated that the optimal number of PCs for the Hyperfast model is between 200 and 500, beyond which performance plateaus or degrades due to feature redundancy.

\begin{table}[t]
\floatconts
  {tab:results-lung}
  {\caption{Binary: Lung Cancer Classification Results}}
  {%
    \sisetup{table-number-alignment=center}
    \setlength{\tabcolsep}{4pt}
    \resizebox{\columnwidth}{!}{%
      \begin{tabular}{>{\centering\arraybackslash}p{4cm}cccccc}
\hline
\textbf{Model (\# of PCs)} & \textbf{Accuracy} & \textbf{AUC} & \textbf{F1-Score} & \textbf{PPV} & \textbf{Sensitivity} & \textbf{Specificity} \\ \hline
\textbf{Hyperfast (200)} & \textbf{0.9945} & \textbf{0.9207} & \textbf{0.9139} & \textbf{1.0000} & \textbf{0.8414} & \textbf{1.0000} \\ 
\textbf{Hyperfast (500)} & \textbf{0.9969} & \textbf{0.9552} & \textbf{0.9531} & \textbf{1.0000} & \textbf{0.9103} & \textbf{1.0000} \\ 
\textbf{Hyperfast (1,000)} & \textbf{0.9955} & \textbf{0.9345} & \textbf{0.9299} & \textbf{1.0000} & \textbf{0.8690} & \textbf{1.0000} \\ 
\textbf{Hyperfast (2,000)} & 0.9859 & 0.7966 & 0.7446 & \textbf{1.0000} & 0.5931 & \textbf{1.0000} \\ \hline
\textbf{XGBoost (200)} & 0.9924 & 0.8930 & 0.8769 & 0.9913 & 0.7862 & 0.9998 \\ 
\textbf{XGBoost (500)} & 0.9940 & 0.9138 & 0.9057 & \textbf{1.0000} & 0.8276 & \textbf{1.0000} \\ 
\textbf{XGBoost (1,000)} & 0.9945 & 0.9207 & 0.9139 & \textbf{1.0000} & 0.8414 & \textbf{1.0000} \\ 
\textbf{XGBoost (2,000)} & \textbf{0.9948} & \textbf{0.9241} & \textbf{0.9179} & \textbf{1.0000} & \textbf{0.8483} & \textbf{1.0000} \\ \hline
\textbf{LightGBM (200)} & 0.9924 & 0.8930 & 0.8769 & 0.9913 & 0.7862 & 0.9998 \\ 
\textbf{LightGBM (500)} & 0.9940 & 0.9138 & 0.9057 & \textbf{1.0000} & 0.8276 & \textbf{1.0000} \\ 
\textbf{LightGBM (1,000)} & 0.9912 & 0.8724 & 0.8538 & \textbf{1.0000} & 0.7448 & \textbf{1.0000} \\ 
\textbf{LightGBM (2,000)} & 0.9905 & 0.8621 & 0.8400 & \textbf{1.0000} & 0.7241 & \textbf{1.0000} \\ \hline
      \end{tabular}
    }
  }
  {}
\end{table}

\tableref{tab:results-lung} shows similar results with Hyperfast model classifying lung cancer with accuracy over 0.9980 until it uses 2,000 PCs for training the model. Similarly, this possible model performance degeneration can be resolved by using our ensemble model where the two boosting algorithms stay relatively robust even when the number of PCs increases. In general, Hyperfast model can play a crucial role in increasing accuracy as well as preventing overfitting when the other two boosting algorithms fail to conduct regularization successfully especially with high-dimensional biomedical data \citep{bonet2024hyperfast}. \citet{yang2022certified} proposed a metric to calculate certified robustness of ensemble model compared to each base models which could be analyzed further in future studies.

\subsection{Multi-class Classification}

We now compare our ensemble model to other baseline machine learning algorithms on multi-class classification tasks, trying to predict all 34 targets, including 33 cancer classes and 1 cancer-negative class. We conduct classification experiments by incrementally increasing the number of PCs from 200 to as many as 15,000 and analyzed the performance using Balanced Accuracy (\figureref{fig:balanced}) and Accuracy (\figureref{fig:accuracy}) as metrics. 

\begin{figure*}[!t] 
\floatconts
  {fig:balanced}
  {\caption{Balanced Accuracy vs.\ number of PCA features (log scale). Markers plotted in black indicate our ensemble model.}}
  {\includegraphics[width=\textwidth]{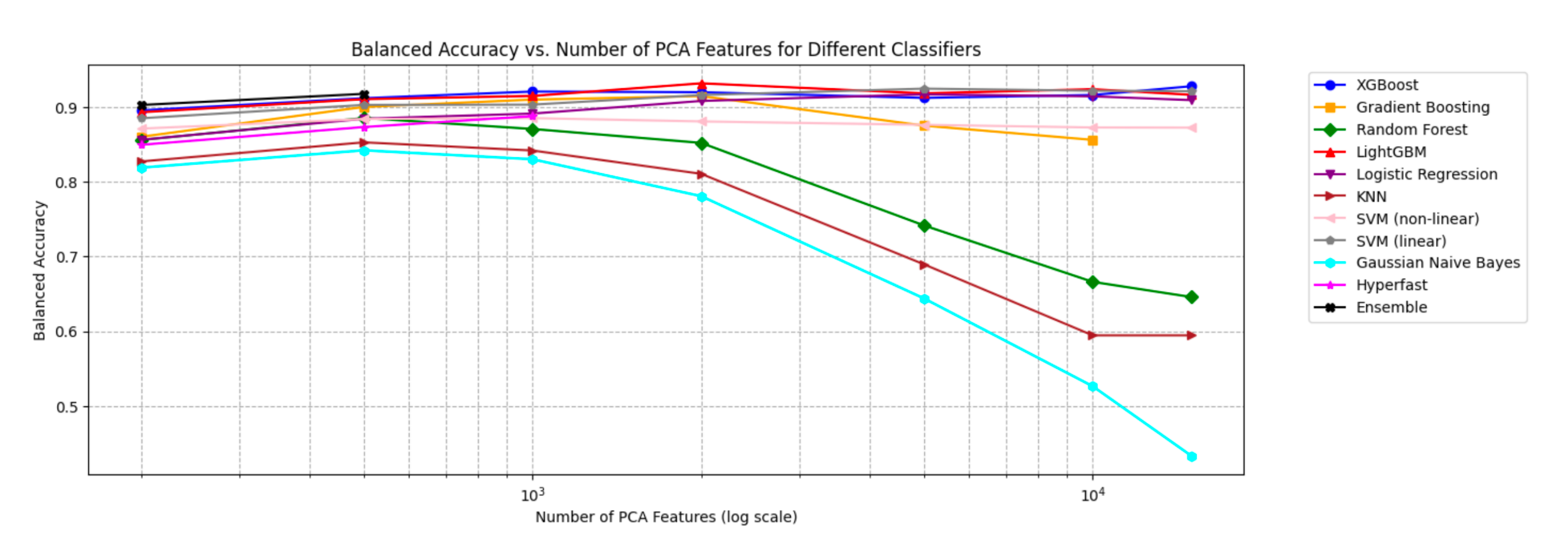}}
\end{figure*}

\begin{figure*}[!t]
\floatconts
  {fig:accuracy}
  {\caption{Accuracy vs.\ number of PCA features (log scale). Markers plotted in black indicate our ensemble model.}}
  {\includegraphics[width=\textwidth]{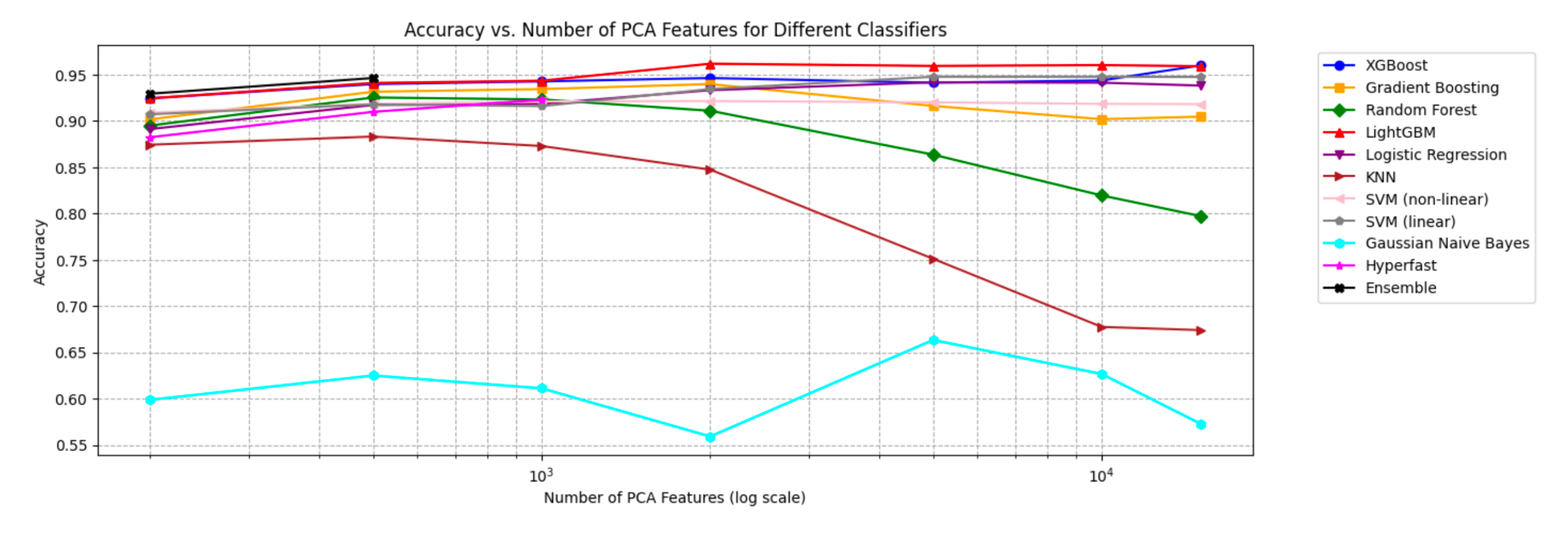}}
\end{figure*}

Both \figureref{fig:balanced} and \figureref{fig:accuracy} demonstrate similar results regarding the rank of the performance of different classifiers. While boosting algorithms such as XGBoost and LightGBM showed accuracy over 0.96 when using over 10,000 PCs, kNN and Gassian Naïve Bayes performed poorly with accuracy less than 0.70 on similar tasks. Additionally, while the accuracy of Random Forest, kNN, and Gaussian Naïve Bayes decreased as the number of PCs increased, the performance of boosting techniques, Logistic Regression, and SVM remained robust even as the number of biomarker features increased up to 15,000. We believe that the possible overfitting occurred in some classification techniques due to failure in regularization (i.e., batch normalization and dropout). 

Due to computing power constraints regarding GPU memory (NVIDIA GeForce RTX 3060 Ti; 24GB Memory), we could only test the performance of our ensemble model until 500 PCs (i.e., 200, 500 PCA features). Out of all models tested, our ensemble model achieved the highest accuracy and balanced accuracy of 0.9464 and 0.9177 respectively when tested using 500 PCs. Also, for experiments using 200 PCs, our ensemble model had the highest accuracy/balanced accuracy of 0.9295 and 0.9028 respectively. These results were slightly higher than the accuracy and balanced accuracy of XGBoost and LightGBM, the two most powerful classification techniques for tabular data.

\subsection{External Validation on Liquid-Biopsy RNA Cohorts}
\label{sec:external}

To assess generalization beyond TCGA tissue biomarkers, we applied the same Hyperfast+XGBoost+LightGBM stack to two external liquid-biopsy RNA cohorts. In both cases we used logCPM normalization, low-expression filtering, top-variance gene selection, standardization, and PCA, with PCA always fit inside each cross-validation training fold to avoid leakage. We report 5-fold stratified cross-validated accuracy (Acc) and balanced accuracy (BAcc); full results appear in Appendix~\ref{apd:external}, Tables~\ref{tab:gse68086} and~\ref{tab:gse71008}.

The first cohort, GSE68086, consists of tumor-educated platelets (TEPs) with seven cancer/other classes (Breast, CRC, GBM, Lung, Pancreatic, Cholangiocarcinoma, Liver; $n=127$, 57{,}736 genes before filtering). After logCPM normalization, low-expression filtering, and restricting to the 5{,}000 most variable genes, we evaluated PCs in $\{25,50,75,100\}$. As shown in Table~\ref{tab:gse68086}, ensemble balanced accuracy ranges from roughly 0.35--0.45 and accuracy from 0.50--0.57 across PC settings. This cohort is small, heterogeneous, and genuinely difficult; absolute performance is modest, but the ensemble is consistently the most reliable model across PCs, typically attaining the highest or near-highest balanced accuracy while Hyperfast alone is competitive only in narrower PC bands.

The second cohort, GSE71008, is a plasma extracellular-vesicle cfRNA dataset for binary Colon vs.\ Healthy classification (imbalance ratio $\approx 2{:}1$, $n=150$ after alignment and filtering). With 50 and 100 PCs derived from 3{,}459 filtered genes, the ensemble attains Acc/BAcc of $(0.74, 0.65)$ and $(0.71, 0.60)$, respectively (Table~\ref{tab:gse71008}), matching or exceeding each base learner on most metrics. These results indicate that, although liquid-biopsy RNA settings are harder than TCGA tissue and yield lower absolute scores, the same pre-trained-plus-ensemble pipeline remains robust and competitive with minimal tuning.

\section{Conclusion}
\label{sec:conclusion}

We predict cancer types from liquid-biopsy biomarkers and show that even a small number of PCA-selected features can support robust performance in both binary and multiclass settings on TCGA tissue biomarkers. For binary classification, leveraging as few as 200 PCs yields $>0.99$ accuracy with the pre-trained Hyperfast model, exceeding the two boosting baselines, and on markedly imbalanced tasks (e.g., LUAD vs.\ non-LUAD) the pre-trained model remains competitive without any task-specific tuning. For multiclass classification over 34 labels, our simple hard-voting ensemble (Hyperfast + XGBoost + LightGBM) attains the highest accuracy/balanced accuracy among all tested methods while using $\leq 500$ PCs, highlighting a favorable accuracy--complexity trade-off. All headline TCGA metrics are obtained on a stratified 75/25 train--test split with PCA fit on the training split only, preventing train--test leakage. We further demonstrate that the same pipeline transfers to two external liquid-biopsy RNA cohorts (tumor-educated platelets and plasma EV cfRNA), where absolute accuracies are modest but the ensemble consistently matches or outperforms its components (Appendix~\ref{apd:external}), supporting robustness beyond TCGA tissue profiles.

Beyond empirical results, we provide a robustness rationale grounded in our analysis. First, we formalize that Hyperfast’s final layer effectively operates in a \emph{prototype} form, per-class averaging in an appropriate feature space, which stabilizes decisions under label skew (\lemmaref{lem:prototype-form}). Under bounded bias, this yields \emph{prior-insensitive} classification margins on imbalanced data, explaining the strong minority recall we observe (\theoremref{thm:HFmargin}). Second, we analyze the ensemble’s hard vote: when base learners exhibit mild error diversity, majority voting reduces minority-class error relative to each constituent model, with guarantees captured by Proposition~7 and strengthened under approximate error independence (\corollaryref{cor:indep}). Finally, we discuss robustness to prior shift (class-frequency changes between train/test), showing conditions under which decision quality degrades gracefully (Proposition~10). Together, these results clarify \emph{why} a pre-trained tabular model plus a simple vote performs reliably under imbalance and distributional drift.

Limitations include our reliance on a single Hyperfast checkpoint \citep{bonet2024hyperfast} and a limited set of external cohorts (one large TCGA tissue panel plus two moderate-size liquid-biopsy studies). Future work should (i) meta-train/fine-tune Hyperfast on institution-specific cohorts to assess broader external validity, (ii) extend the theory to calibrated decisions and cost-sensitive risk under extreme imbalance, and (iii) evaluate the framework on multi-domain, non-tabular modalities (e.g., imaging, videos) where pre-training and simple ensembling may likewise offer robustness with minimal tuning.

\acks{This research was supported by the MSIT (Ministry of Science and ICT), Korea, under the ITRC (Information Technology Research Center) support program (IITP-2024-2020-0-01789), and the Artificial Intelligence Convergence Innovation Human Resources Development program (IITP-2024-RS-2023-00254592) supervised by the IITP (Institute for Information \& Communications Technology Planning \& Evaluation).}

\bibliography{jmlr-sample}

\appendix

\section{Robustness to Class Imbalance: Hyperfast and Hard-Vote Ensemble}
\label{apd:first}

\subsection{Setting and Notation}
Let $\mathcal{X}\subset\mathbb{R}^d$ and $\mathcal{Y}=\{1,\dots,C\}$. For class $c$, data are i.i.d.\ from $P_c$ with prior $\pi_c=\Prb(Y=c)$ (possibly imbalanced). A \emph{support} $S=\bigcup_{c=1}^C S_c$ is class-balanced with
$S_c=\{(x^{(c)}_j,c)\}_{j=1}^{k}$ for all $c$. The \emph{query} $Q$ is an i.i.d.\ sample used for evaluation.
We denote by $\Phi_S:\mathcal{X}\to\mathbb{R}^p$ the feature map produced by the Hyperfast initial transformation (random features followed by PCA fitted on $S$), and by $\langle\cdot,\cdot\rangle$ and $\|\cdot\|$ the Euclidean inner product and norm.

\paragraph{Hyperfast last-layer construction (as implemented).}
The final (classification) layer forms one row per class via \emph{per-class averaging} of support representations, plus a small residual from the hypernetwork module, and a class-dependent bias driven by the NN-bias head; cf.\ the architecture description in the main text.

\begin{definition}[Hyperfast-specific assumptions]\label{def:Hassumptions}
We assume:
\begin{enumerate}
\item[(H1)] \textbf{Bounded transformed features.} There exists $B>0$ such that $\|\Phi_S(x)\|\le B$ almost surely for all $x$ from the (support/query) distribution conditional on $S$.
\item[(H2)] \textbf{Prototype+\;residual final layer.} For class $c$,
\begin{equation}
\begin{aligned}
w_c &= \mu_c(S) + r_c(S),\\
\mu_c(S) &:= \frac{1}{k}\sum_{(x,c)\in S_c}\Phi_S(x),\\
\|r_c(S)\| &\le \rho.
\end{aligned}
\end{equation}
and the bias $b_c(S)$ depends only on $S$ (episode-local statistics).
\item[(H3)] \textbf{Prior-neutral episode.} $S$ is class-balanced; $b_c(S)$ does not encode global priors $\{\pi_c\}$.
\end{enumerate}
\end{definition}

Hyperfast predicts
\[
\hat y(x;S)=\arg\max_{c\in\mathcal{Y}} \; \langle w_c,\Phi_S(x)\rangle + b_c(S).
\]

\subsection{Hyperfast as a Prototype Classifier and Prior-Independent Margin}
\begin{lemma}[Prototype decision form]\label{lem:prototype-form}

Under \definitionref{def:Hassumptions}, with $\tilde b_c(S):=b_c(S)+\langle r_c(S),\Phi_S(x)\rangle$, we have
\begin{align}
\hat y(x;S)
&= \operatorname*{arg\,max}_{c}\;
   \big\langle \mu_c(S),\,\Phi_S(x)\big\rangle \nonumber\\
&\qquad + \tilde b_c(S), \label{eq:pred}\\[2pt]
\text{and}\quad
\big|\tilde b_{c'}(S)-\tilde b_c(S)\big|
&\le 2\rho B .
\end{align}
\begin{proof}
Substitute $w_c=\mu_c(S)+r_c(S)$ into the score and use (H1): $|\langle r_c(S),\Phi_S(x)\rangle|\le \|r_c(S)\|\,\|\Phi_S(x)\|\le \rho B$.
\end{proof}
\end{lemma}

Define population prototypes 

$\mu_c:=\E\!\left[\Phi_S(X)\mid Y=c\right]\in\mathbb{R}^p$ and the separation
\[
\Delta \;:=\; \min_{c\neq c'} \|\mu_c-\mu_{c'}\| \;>\;0.
\]

\begin{lemma}\label{lem:concentration}
\textbf{(Concentration of empirical prototypes).}

Under (H1), for any $\delta\in(0,1)$,
\[
\Prb\!\left(\max_{c}\,\|\mu_c(S)-\mu_c\| \;\le\; B\sqrt{\frac{2\log(2C/\delta)}{k}}\right) \;\ge\; 1-\delta.
\]
\begin{proof}
Apply a vector Hoeffding/Bernstein inequality to each class with bounded summands by (H1) and union bound over $C$ classes.
\end{proof}
\end{lemma}

\begin{theorem}\label{thm:HFmargin}
\textbf{(Prior-independent Hyperfast margin and error).}

Assume \definitionref{def:Hassumptions} and $\Delta>0$. Let
\[
\epsilon_k \;:=\; B\sqrt{\frac{2\log(2C/\delta)}{k}} + \rho,
\qquad 
\gamma \;:=\; \tfrac12\,\Delta - \epsilon_k.
\]
If $\gamma>0$, then with probability at least $1-\delta$ over $S$,
\[
\frac{1}{C}\sum_{c=1}^C \Prb_{X\sim P_c}\!\big(\hat y(X;S)\neq c\big)
\;\le\; C\,\exp\!\Big(-\frac{\gamma^2}{2B^2}\Big),
\]
and this bound is \emph{independent of the class priors} $\{\pi_c\}$.

\begin{proof}
By \lemmaref{lem:prototype-form}, misclassifying $c$ as $c'$ requires that the sum of (i) prototype estimation error (bounded by \lemmaref{lem:concentration}) and (ii) residual/bias perturbation (bounded by $\rho B$) overcomes half the inter-prototype distance. The resulting effective margin is at least $\tfrac12\|\mu_{c'}-\mu_c\|-\epsilon_k$. A bounded linear-margin tail bound (in a $B$-ball) yields $\Prb_{P_c}(c\to c')\le \exp(-\gamma^2/(2B^2))$ when $\gamma>0$; a union bound over $c'\neq c$ and an average over classes gives the result. No $\pi_c$ appears.
\end{proof}
\end{theorem}

\paragraph{Consequences.}
(i) Robustness to imbalance arises because the final decision is \emph{prototype-based} with equal $k$ per class, so margins and rates do not depend on priors.  
(ii) Minority-class accuracy scales with $k$ (support size per class), not with the minority prior, protecting recall for rare classes.  
(iii) Keeping $\rho$ small (via weight decay/early stopping) preserves the effective margin $\gamma$.

\subsection{Three-Model Hard-Vote Ensemble: Class-wise Error Bounds}
Let $j\in\{H,L,X\}$ index Hyperfast, LightGBM, and XGBoost. For class $c$, define the base error
\begin{align}
e_j(c) &= \Prb_{X\sim P_c}\big(\hat y_j(X)\neq c\big), \notag\\
I^{(j)}_c(X) &:= \mathbf{1}\{\hat y_j(X)\neq c\}.
\end{align}
The majority-vote (hard-vote) ensemble errs on class $c$ iff at least two bases err:
\[
e_{\text{ens}}(c) \;=\; \Prb\big(I^{(H)}_c+I^{(L)}_c+I^{(X)}_c\ge 2\big).
\]

\begin{proposition}\label{prop:identity}
\textbf{(Exact decomposition for three voters).}\\
Let $p_{jk}(c):=\Prb\!\big(I^{(j)}_c=1,\,I^{(k)}_c=1\big)$ and $p_{123}(c):=\Prb(I^{(H)}_c=I^{(L)}_c=I^{(X)}_c=1)$. Then
\begin{equation}
e_{\text{ens}}(c) \;=\; p_{HL}(c)+p_{HX}(c)+p_{LX}(c)\;-\;2\,p_{123}(c).
\label{eq:hv-identity}
\end{equation}
\begin{proof}
“At least two wrong” equals “exactly two wrong” plus “all three wrong”. Each $p_{jk}$ counts the $(j,k)$-only event once and the triple-wrong event once; summing the three $p_{jk}$ thus counts the triple-wrong event thrice, so subtract two copies.
\end{proof}
\end{proposition}

\begin{corollary}[A simple upper bound]\label{cor:upper}

Dropping the nonnegative triple term in \equationref{eq:hv-identity} gives
\[
e_{\text{ens}}(c) \;\le\; p_{HL}(c)+p_{HX}(c)+p_{LX}(c).
\]
\end{corollary}

Write $p_{jk}(c)=e_j(c)e_k(c)+\mathrm{Cov}(I^{(j)}_c,I^{(k)}_c)$. Under a pairwise covariance cap,
\begin{equation}\label{eq:covcap}
\mathrm{Cov}(I^{(j)}_c,I^{(k)}_c) \;\le\; \kappa_c \qquad \text{for all pairs }(j,k),
\end{equation}
we obtain the following class-wise bound.

\begin{proposition}\label{prop:bounded}
\textbf{(Ensemble error with bounded dependence).}

Under \equationref{eq:covcap},
\[
e_{\text{ens}}(c) 
\;\le\; e_H(c)e_L(c)+e_H(c)e_X(c)+e_L(c)e_X(c) \;+\; 3\,\kappa_c.
\]
\end{proposition}

\begin{corollary}[Independence special case]\label{cor:indep}

If the indicators $\{I^{(j)}_c\}$ are independent for class $c$, then
\[
e_{\text{ens}}(c)
= e_H e_L + e_H e_X + e_L e_X - 2 e_H e_L e_X.
\]
\end{corollary}

\begin{corollary}\label{cor:symmetric}
\textbf{(Symmetric sufficient improvement).}

If $e_H(c)=e_L(c)=e_X(c)=\varepsilon<\tfrac12$ and \equationref{eq:covcap} holds, then
$e_{\text{ens}}(c)\le 3\varepsilon^2+3\kappa_c$, hence $e_{\text{ens}}(c)<\varepsilon$ whenever $\kappa_c<(\varepsilon-3\varepsilon^2)/3$.
\end{corollary}

\paragraph{Balanced error and priors.}
All bounds above are \emph{class-wise}; the balanced ensemble error $\frac{1}{C}\sum_c e_{\text{ens}}(c)$ is therefore \emph{independent of the priors} $\{\pi_c\}$ by construction.

\subsection{Prior-Shift Resilience with One Prior-Invariant Voter}
Let $p_j^{\mathrm{flip}}(c)$ be the probability (over $X\sim P_c$) that model $j$ \emph{changes} its class-$c$ decision after a class-prior shift (development $\to$ deployment). By \theoremref{thm:HFmargin}, Hyperfast is prior-insensitive under balanced episodic construction, so set $p_H^{\mathrm{flip}}(c)=0$.

\begin{proposition}\label{prop:flip}
\textbf{(Flip probability of majority vote).}\\
For any dependence structure,
\begin{align}
\Prb\big(\text{maj.\ vote flips on }c\big)
&\le \Prb\big(\text{L\ \&\ X both flip on }c\big) \notag\\
&\le \min\{p_L^{\mathrm{flip}}(c),\,p_X^{\mathrm{flip}}(c)\}.
\end{align}

Under independence of flips, $\Prb(\text{flip on }c)\le p_L^{\mathrm{flip}}(c)\,p_X^{\mathrm{flip}}(c)$.\\

\begin{proof}
With $H$ invariant, the majority outcome changes only if both remaining voters change simultaneously. The bounds follow by basic probability inequalities; the independence case multiplies probabilities.
\end{proof}
\end{proposition}

\paragraph{Implications.}
(i) Including Hyperfast \emph{damps} the ensemble’s sensitivity to prior shifts, especially for minority classes.  
(ii) Architectural diversity (prototype-based Hyperfast vs.\ tree-based learners) empirically reduces the pairwise joint-error terms in Proposition 7, tightening the bound.

\section{External Validation on Liquid-Biopsy RNA Cohorts}
\label{apd:external}

\subsection{Preprocessing and Evaluation Protocol}

For both external RNA cohorts we followed a consistent preprocessing and evaluation protocol designed to avoid data leakage and to mirror the TCGA pipeline.

For RNA-seq-style data we first computed log-counts-per-million (logCPM), then removed low-expression genes using a CPM $>1$ threshold in at least 10\% of samples. Among the remaining genes we selected the top-variance features (we explored 2{,}000--5{,}000 genes and settled on 5{,}000 for GSE68086 and 3{,}459 for GSE71008). All features were standardized before PCA. Crucially, PCA was always fit \emph{inside} each cross-validation training fold (or on the training split only) and then applied to the corresponding validation/test partition, so that no information from held-out samples leaked into the feature construction.

For model evaluation we used 5-fold stratified cross-validation to preserve class proportions within each fold. For each number of PCs we trained Hyperfast, XGBoost, and LightGBM on the training folds and evaluated accuracy (Acc) and balanced accuracy (BAcc) on the held-out folds. The ensemble (ENS) prediction is the hard vote over the three base learners, as in Equation~\eqref{eq:mv}.

\subsection{GSE68086: Tumor-Educated Platelets (Multiclass)}

GSE68086 contains tumor-educated platelet RNA profiles from multiple cancer types and controls (Breast, Cholangiocarcinoma, Colorectal cancer, Glioblastoma, Lung, Liver, and other classes), with $n=127$ samples and 57{,}736 genes before filtering. After preprocessing and restricting to the 5{,}000 most variable genes, we evaluated PCs in $\{25,50,75,100\}$. Table~\ref{tab:gse68086} reports mean 5-fold cross-validated metrics.

\begin{table}[H]
\floatconts
  {tab:gse68086}
  {\caption{External validation on GSE68086 (tumor-educated platelets; multiclass). Mean 5-fold cross-validated accuracy/balanced accuracy (Acc/BAcc) for Hyperfast (HF), XGBoost (XGB), LightGBM (LGB), and the ensemble (ENS).}}
  {%
    \setlength{\tabcolsep}{2pt}
    \scriptsize
    \begin{tabular}{c|c|c|c|c}
      \toprule
      \textbf{PCs} & \textbf{HF} & \textbf{XGB} & \textbf{LGB} & \textbf{ENS} \\
      \midrule
      25  & 0.487 / 0.370 & 0.552 / 0.472 & 0.504 / 0.404 & 0.559 / 0.454 \\
      50  & 0.543 / 0.388 & 0.527 / 0.417 & 0.543 / 0.409 & 0.567 / 0.428 \\
      75  & 0.380 / 0.273 & 0.497 / 0.356 & 0.480 / 0.337 & 0.504 / 0.353 \\
      100 & 0.462 / 0.363 & 0.505 / 0.374 & 0.489 / 0.351 & 0.528 / 0.384 \\
      \bottomrule
    \end{tabular}
  }%
\end{table}

Across all PC settings the ensemble achieves the most reliable performance: balanced accuracy is typically highest or near-highest, with overall accuracy in the 0.50--0.57 range. Hyperfast alone is competitive in some PC regimes (e.g., 25 and 50 PCs) but degrades more sharply for others, consistent with our ensemble rationale in weak-signal, heterogeneous multiclass settings.

\subsection{GSE71008: Plasma EV cfRNA (Binary Colon vs.\ Healthy)}

GSE71008 consists of plasma extracellular-vesicle cfRNA profiles for Colon cancer vs.\ Healthy controls, with $n=150$ samples (100 Colon, 50 Healthy) and 3{,}460 genes before filtering. After low-expression filtering we retained 3{,}459 genes, then evaluated 50 and 100 PCs with the same 5-fold stratified CV protocol. Results are in Table~\ref{tab:gse71008}.

\begin{table}[t]
\floatconts
  {tab:gse71008}
  {\caption{External validation on GSE71008 (plasma EV cfRNA; binary Colon vs.\ Healthy). Mean 5-fold cross-validated accuracy/balanced accuracy (Acc/BAcc).}}
  {%
    \setlength{\tabcolsep}{2pt}
    \scriptsize
    \begin{tabular}{c|c|c|c|c}
      \toprule
      \textbf{PCs} & \textbf{HF} & \textbf{XGB} & \textbf{LGB} & \textbf{ENS} \\
      \midrule
      50  & 0.727 / 0.640 & 0.727 / 0.645 & 0.707 / 0.620 & 0.740 / 0.650 \\
      100 & 0.747 / 0.685 & 0.720 / 0.625 & 0.700 / 0.570 & 0.713 / 0.600 \\
      \bottomrule
    \end{tabular}
  }%
\end{table}

Here the ensemble again achieves the most reliable performance, reproducing a balanced accuracy of $\sim$0.65 with a compact and fast pipeline. While Hyperfast occasionally attains slightly higher balanced accuracy at a single PC setting, the ensemble is competitive or superior across both PC choices, reinforcing the value of combining diverse inductive biases when signal-to-noise is limited.

\end{document}